\documentclass[10pt]{article}
\PassOptionsToPackage{table}{xcolor}
\usepackage{adbpreprint}
\usepackage[authoryear,round]{natbib}
\setcitestyle{authoryear,round,citesep={;},aysep={,},yysep={;}}
\usepackage{amsmath,amssymb}
\usepackage{booktabs}
\usepackage{float}
\usepackage{graphicx}
\usepackage{longtable}
\usepackage{placeins}
\usepackage{tabularx}
\usepackage{enumitem}
\newtcolorbox{keyfindings}{
  colback=adblink!5,
  colframe=adblink!75,
  fonttitle=\sffamily\bfseries,
  title=Key Findings,
  arc=2mm,
  boxrule=1pt,
  before upper={\setlist[itemize]{itemsep=0.5\baselineskip}},
}
\usepackage[colorlinks=true,linkcolor=adblink,citecolor=adblink,urlcolor=adblink]{hyperref}
\usepackage{url}

\usepackage{amsmath,amsfonts,bm}

\def\eqref#1{equation~\ref{#1}}

\def\1{\bm{1}}

\DeclareMathAlphabet{\mathsfit}{\encodingdefault}{\sfdefault}{m}{sl}
\SetMathAlphabet{\mathsfit}{bold}{\encodingdefault}{\sfdefault}{bx}{n}

\hypersetup{
  pdftitle={Long-Horizon Analog Design Bench: Benchmarking Agents on Hours-Long Analog and Mixed-Signal Circuit Design Tasks},
  pdfauthor={Analog Design Bench Team},
}

\title{Long-Horizon Analog Design Bench:\\
Benchmarking Agents on Hours-Long\\
Analog and Mixed-Signal Circuit Design Tasks}
\byline{Analog Design Bench Team}
\abstracttext{Coding agents now sustain hours-long, tool-driven loops, yet their ability to carry long-horizon analog and mixed-signal circuits to electrical specification remains unmeasured.
We introduce \emph{Analog Design Bench}, a long-horizon agentic benchmark of 50 transistor-level design tasks contributed by 17 chip designers.
Agents work with an open-source simulator, while an isolated verifier evaluates the submitted circuit using specification-based electrical tests.
We evaluate 15 agent configurations across 2,250 two-hour attempts and observe full-specification pass rates from 8.0\% to 78.0\%.
Coding-benchmark performance correlates with analog results but leaves much of the performance spread unexplained.
Our failure analysis shows that most unsuccessful submissions have no recorded legality rejection but fail electrical acceptance, identifying electrical closure as the dominant endpoint challenge.
We test time, reasoning effort, agent harness, and supplied design knowledge as interventions.
Longer budgets and higher reasoning effort improve performance, while general skill documents provide little benefit and sometimes reduce performance.
Supplying a task-matched reference topology, an idealized form of circuit-IP retrieval, raises DeepSeek V4 Pro by 18.7 percentage points and mainly accelerates GPT-5.6 Sol.
The released benchmark, trajectories, and analysis tools provide a testbed for developing long-horizon agents for physics-grounded analog and mixed-signal integrated circuit design.}
\metadata{Correspondence}{Zhishuai Zhang at \texttt{zzhishuai@ethz.ch}}
\metadata{Project Page}{\url{https://analog-design-bench.tokenzhang.com}}
\metadata{Repository}{\url{https://github.com/Arcadia-1/analog-design-bench}}

\makeatletter
\newenvironment{inplacefigures}{%
  \renewcommand{\suppressfloats}[1][]{}%
  \let\adb@figure\figure
  \def\figure[##1]{\adb@figure[H]}}{}
\makeatother

\begin{document}
\maketitle
\begin{inplacefigures}
\suppressfloats[t]
\begin{figure}[t]
  \centering
  \includegraphics[width=0.9\linewidth]{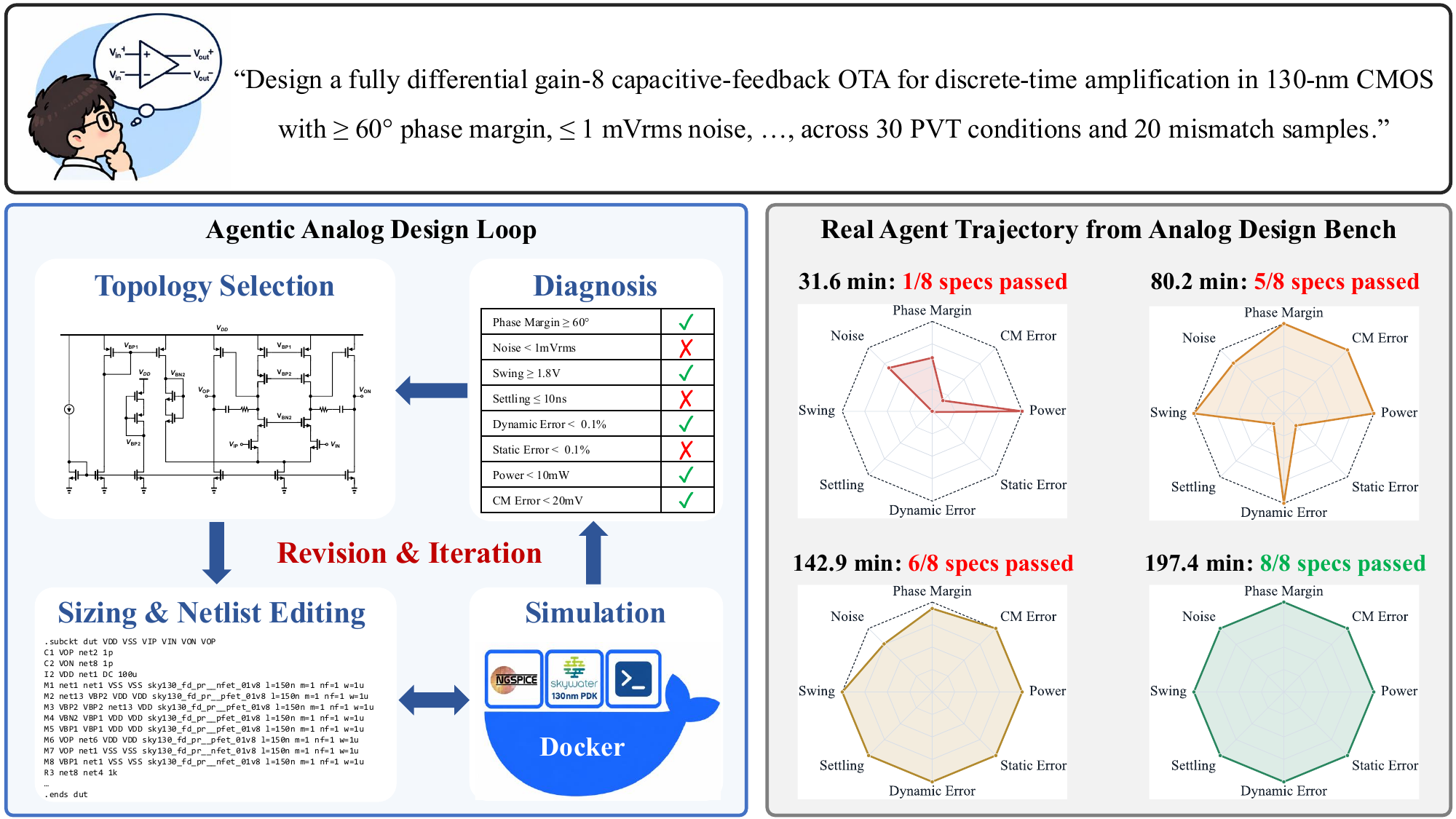}
  \caption{Agentic analog design loop and a real Analog Design Bench trajectory. Agents iteratively select topologies, edit and size circuits, simulate, diagnose, and revise toward electrical closure. Right: GPT-5.6 Sol [max] reaches its first full pass at 197.4 min. Radar plots show eight representative specifications from the task's 15 scoring gates.}
  \label{fig:headline}
\end{figure}
\end{inplacefigures}
\section{Introduction}

Coding agents now sustain long-horizon workflows in software engineering and terminal environments through iterative editing, tool use, and execution feedback~\citep{luo2025agentsurvey,li2026issueresolution,yang2024sweagent,merrill2026terminalbench}.
Similar agentic workflows are beginning to spread into electronic design automation~\citep{pan2025llmeda,zang2025agenticeda}.
Front-end analog design starts from an electrical specification.
Engineers select and adapt a circuit topology, size devices, construct diagnostic testbenches, and iterate on simulator feedback until coupled functionality, performance, and robustness requirements hold across operating conditions.
This process requires extensive expertise and manual tuning because a syntactically valid netlist can still implement the wrong function or miss a performance limit~\citep{razavi2001design,gray2009analysis}.
This reliance on expert iteration has long motivated analog design automation~\citep{gielen2000cad,lyu2018weibo,gao2025analoggenie}.
These activities fit a tool-using agent workflow, yet the ability of general-purpose coding agents to complete them end to end remains underexplored.
Figure~\ref{fig:headline} summarizes this iterative workflow and a real benchmark trajectory.

LLM-based analog-design systems have progressed from circuit generation toward simulator-in-the-loop, multi-agent, and memory-augmented workflows~\citep{lai2025analogcoder,lai2026analogcoderpro,liu2024ampagent,shen2026atelier,bao2026analogagent,DBLP:journals/corr/abs-2604-23195}.
These studies establish feasibility, while their evaluations use smaller suites, specialized agents, or protocols that do not jointly provide original expert-contributed tasks, multi-hour autonomous simulator use, and isolated specification-based verification (Table~\ref{tab:benchmark-positioning}).
Analog Design Bench targets this missing combination.

More than twenty analog designers proposed over 100 problems.
Multi-stage author, domain, and meta review of specifications, reference results, testbenches, shortcut risks, and trial trajectories retained 50 tasks from 17 experts.
To our knowledge, Analog Design Bench is the first analog benchmark to combine original expert-contributed tasks, multi-hour autonomous simulator use, and isolated specification-based verification.

We conduct a comprehensive evaluation of 15 agent configurations, jointly defined by model, reasoning effort, and harness, over 2,250 two-hour attempts.
Claude Fable 5 leads at 78.0\%, followed by Claude Opus 5, GPT-5.6 Sol, and GPT-5.5, while the remaining eleven configurations score below 50\%.
All four configurations above 50\% use proprietary frontier models.

Three findings stand out.
First, coding and analog rankings are correlated, yet coding scores leave much of the analog-performance spread unexplained~\citep{huang2026deepswe}.
Second, 98.3\% of non-passing attempts have no recorded legality rejection but fail electrical acceptance.
Third, longer runs and greater reasoning effort improve pass rates; harness differences narrow with time, general-purpose skills have small or inconsistent effects, and a task-matched reference topology lifts a weaker model substantially while mainly accelerating a stronger one.

The paper makes three contributions.
First, we introduce 50 expert-contributed transistor-level tasks with an open-source toolchain, an isolated verifier, and specification-based electrical tests.
Second, we systematically evaluate 15 agent configurations with three rollouts per task, quantifying capability differences, reliability, and resource use.
Third, we identify the factors that shape performance through failure analysis, test-time interventions, supplied design knowledge, and circuit-level trajectories.
Together, the benchmark and findings provide a basis for developing more capable long-horizon analog-design agents.%
\section{Related Work}
\label{sec:related}

\paragraph{Agentic and hardware benchmarks.}
SWE-bench evaluates repository patches with executable tests \citep{jimenez2024swebench}; DeepSWE uses original software-engineering tasks and hand-written verifiers \citep{huang2026deepswe}.
Terminal-Bench evaluates agents on realistic tasks in containerized command-line environments \citep{merrill2026terminalbench}.
Verifier coverage \citep{liu2023evalplus} and the agent harness \citep{yang2024sweagent,wang2024openhands,xia2024agentless} both affect the measurement.
VerilogEval~\citep{liu2023verilogeval} and RTLLM~\citep{lu2024rtllm} evaluate RTL generation.
CVDP (Comprehensive Verilog Design Problems) includes 166 agentic and 617 non-agentic tasks for RTL design, verification, and comprehension \citep{pinckney2025cvdp}.

\paragraph{Analog design automation.}
AutoCkt uses reinforcement learning for simulator-guided sizing \citep{settaluri2020autockt}; related approaches include Bayesian optimization \citep{lyu2018weibo} and graph-based policies \citep{wang2020gcnrl}, with AnalogGym providing executable sizing tasks \citep{li2024analoggym}.
Learned topology generation extends the search beyond device parameters \citep{dong2023cktgnn,chang2024lamagic,gao2025analoggenie}.
AnalogCoder and AnalogCoder-Pro iterate generation and simulation \citep{lai2025analogcoder,lai2026analogcoderpro}; AmpAgent, Atelier, and AnalogAgent add specialized reasoning, roles, or memory \citep{liu2024ampagent,shen2026atelier,bao2026analogagent}.
Analog Design Bench combines the four properties of Table~\ref{tab:benchmark-positioning} that no prior analog benchmark offers together: original expert-contributed tasks, \emph{long-horizon} runs of two to six hours, autonomous simulator use, and an isolated verifier.
Appendix~\ref{app:related} gives the extended survey.

\begin{table}[t]
  \definecolor{tableyesgreen}{HTML}{16835A}
  \definecolor{tablenored}{HTML}{C43C45}
  \hypersetup{colorlinks=true,citecolor=adblink}
  \let\tablecitep\citep
  \renewcommand{\citep}[1]{{\color{adblink}\tablecitep{#1}}}
  \caption{Related work and positioning. Among the listed analog benchmarks, Analog Design Bench uniquely combines original expert-contributed tasks, long-horizon tool use, autonomous simulation, and isolated verification.}
  \label{tab:benchmark-positioning}
  \vspace{2pt}
  \centering\fontsize{7.5}{9}\selectfont
  \newcommand{\PositioningCell}[2][l]{%
    \begin{tabular}[t]{@{}#1@{}}\let\newline\\#2\end{tabular}}
  \newcommand{\PositioningYes}{{\color{tableyesgreen}\ensuremath{\bm{\checkmark}}}}
  \newcommand{\PositioningNo}{{\color{tablenored}\ensuremath{\bm{\times}}}}
  \newcommand{\PositioningYesBold}{\PositioningYes}
  \setlength{\tabcolsep}{1.5pt}
  \begin{tabular*}{0.9\linewidth}{@{\extracolsep{\fill}}llcccccc@{}}
    \toprule
    Work & Task & Scoring & \# Tasks & \PositioningCell[c]{Original\newline tasks} & \PositioningCell[c]{Long\newline horizon\textsuperscript{*}} & \PositioningCell[c]{Tool-use\newline autonomy\textsuperscript{\ensuremath{\dagger}}} & \PositioningCell[c]{Isolated\newline verification} \\
    \midrule
    \multicolumn{8}{@{}l@{}}{\textbf{General Coding}} \\
    Terminal-Bench~\citep{merrill2026terminalbench} & Terminal tasks & AT & 89 & \PositioningYes & \PositioningYes & \PositioningYes & \PositioningNo \\
    DeepSWE~\citep{huang2026deepswe} & Repository changes & AT & 113 & \PositioningYes & \PositioningYes & \PositioningYes & \PositioningYes \\
    \midrule
    \multicolumn{8}{@{}l@{}}{\textbf{Digital Circuit}} \\
    VerilogEval~\citep{liu2023verilogeval} & RTL design & AT & 156 & \PositioningNo & \PositioningNo & \PositioningNo & \PositioningNo \\
    RTLLM~\citep{lu2024rtllm} & RTL design & AT & 50 & \PositioningYes & \PositioningNo & \PositioningNo & \PositioningNo \\
    CVDP~\citep{pinckney2025cvdp} & RTL design & AT & 783 & \PositioningYes & partial & partial & partial \\
    \midrule
    \multicolumn{8}{@{}l@{}}{\textbf{Analog Circuit}} \\
    AnalogCoder~\citep{lai2025analogcoder} & Circuit Design & AT & 24 & \PositioningYes & \PositioningNo & \PositioningNo & \PositioningNo \\
    AnalogCoder-Pro~\citep{lai2026analogcoderpro} & Circuit Design & AT & 40 & \PositioningYes & \PositioningNo & \PositioningNo & \PositioningNo \\
    AnalogAgent~\citep{bao2026analogagent} & Circuit Design & AT & 30 & \PositioningNo & \PositioningYes & \PositioningNo & \PositioningNo \\
    Razavi-Bench~\citep{zhang2026razavibench} & Circuit Analysis & RJ & 50 & \PositioningYes & \PositioningNo & \PositioningNo & \PositioningNo \\
    \textbf{Analog Design Bench (this work)} & \textbf{Circuit Design} & \textbf{AT} & \textbf{50} & \PositioningYesBold & \PositioningYesBold & \PositioningYesBold & \PositioningYesBold \\
    \bottomrule
  \end{tabular*}
  \par\smallskip
  \begin{minipage}{0.9\linewidth}
    \fontsize{7.5}{9}\selectfont
    AT: Automated Test; RJ: Rubric Judge.\par
    \textsuperscript{*}At least 10 stateful tool-feedback rounds.
    \textsuperscript{\ensuremath{\dagger}}Agent chooses when and how to invoke task-relevant tools; fixed framework-run simulations do not count.
  \end{minipage}
\end{table}
\vspace{-2pt}%
\section{Benchmark Design}

This section covers task sourcing, task format, and verification (Sections~\ref{sec:suite}--\ref{sec:verification}).

\subsection{Task sourcing and coverage}
\label{sec:suite}

We sought realistic, diverse tasks that distinguish current agents, proposed by contributors with experience in circuit design, tape-out, and silicon validation across sub-domains.
More than twenty designers proposed 100+ tasks; author, domain reviewer, and meta-reviewer checks of targets, references, testbenches, and shortcuts, together with trial runs on six agent configurations (Appendix~\ref{app:evidence-limits}) to exclude easily solved tasks, yielded 50 tasks from 17 experts.
Each retained task has an independently verified reference result, providing evidence that its contract is achievable without prescribing the agent's topology.

The suite spans six families: power management and references (11), general-purpose op amps and OTAs (9), signal-chain amplifiers and active filters (9), data conversion and sampling (9), RF/timing/high-speed circuits (9), and interfaces and drivers (3).
The data-conversion and sampling family additionally exercises mixed-signal behavior such as switching, timing, and quantization.
Problems range from a five-transistor OTA to asynchronous SAR ADCs and a high-speed CML driver.
Appendix~\ref{app:tasks} lists concise design objectives and summarizes contributor and measurement coverage across all six circuit families.

\begin{figure}[t]
  \centering
  \includegraphics[width=0.9\linewidth]{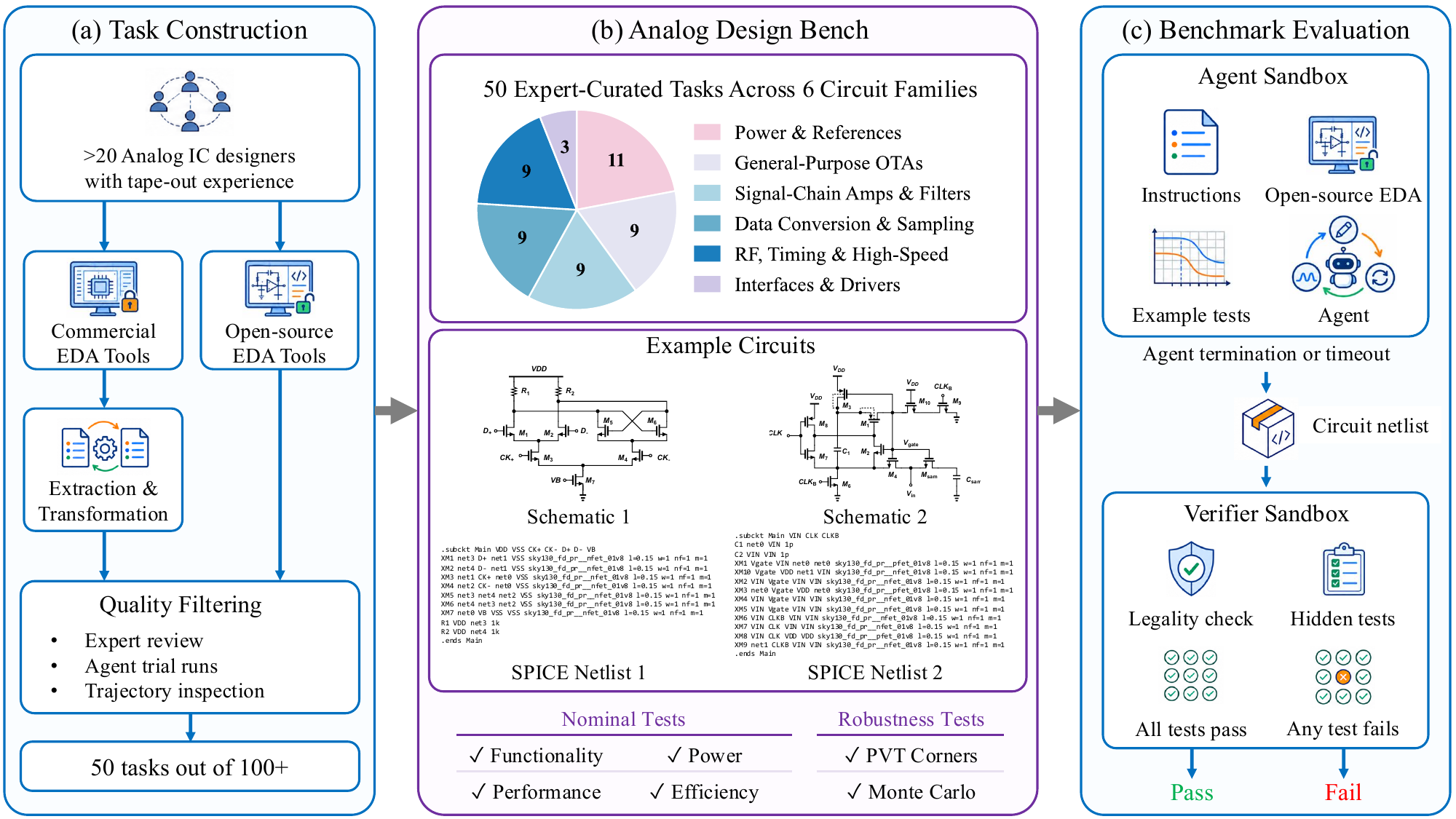}
  \caption{Benchmark construction and evaluation.
  Left: task construction and screening, from 100+ candidates by 20+ designers to 50 retained tasks by 17 contributors.
  Center: task counts across six circuit families.
  Right: the agent edits and simulates in one sandbox; only its declared circuit crosses into an independent verifier sandbox for legality checks and specification-based electrical grading.
  Hidden grading results remain inaccessible to the agent during design attempts.}
  \label{fig:boundary}
\end{figure}

\subsection{Task format and agent environment}
\label{sec:format}

After review, each problem is converted into a uniform format with an electrical contract, interface, starter files, netlist guide, and example testbenches for syntax reference.
We use the open-source SKY130 PDK and ngspice for reproducible evaluation without proprietary data or licenses~\citep{skywater2020pdk,ngspice2026manual}.
We focus on schematic-level design, which captures the core work of front-end analog circuit designers: selecting circuit topologies, sizing devices, and closing electrical specifications through simulation.
Layout and physical implementation form a subsequent design stage and remain outside the benchmark.
The agent writes the device under test (DUT) and diagnostic testbenches, edits files, runs simulations, and refines the design from measured results; a run ends when the agent stops on its own or when an undisclosed wall-clock limit expires.
Each task is bounded so that a single simulation takes at most roughly three minutes, keeping tool calls short and allowing many design iterations within a run.

Adversarial review exposed \emph{reward hacking}, including attempts to alter PDK temperature coefficients.
We therefore strengthened the legality checks and isolated verifier, allowing only the submitted circuit to cross the sandbox boundary while protecting the PDK and tests.

\subsection{Verification and scoring}
\label{sec:verification}
The implementation of the grading benches remains hidden, while the specifications they evaluate are fully disclosed.
This separation reduces overfitting to visible test implementations without introducing undisclosed requirements.
Only the circuit submitted by the agent is judged.
Upon receiving the declared circuit, the isolated verifier checks the interface, enforces permitted device primitives, rejects prohibited idealized shortcuts, and then executes the hidden electrical benches.
These benches implement \emph{specification-based machine evaluation}: each specification item is compiled into a \emph{gate} that compares a named measurement against a prescribed threshold or validity condition.
Each named scoring gate may aggregate many operating conditions.
Expanding the metrics over their discrete prescribed conditions yields 11 to 2,083 electrical acceptance checks per task, excluding raw waveform samples and continuous-sweep discretization points.
A run passes only if every acceptance check passes; as a diagnostic of partial progress, we additionally record gate reward as the proportion of scoring-gate weight earned by passed gates, using task-declared weights when specified and equal weights otherwise.%
\section{Main Evaluation}
\label{sec:cohort}

\begin{figure}[t]
  \centering
  \includegraphics[width=\linewidth]{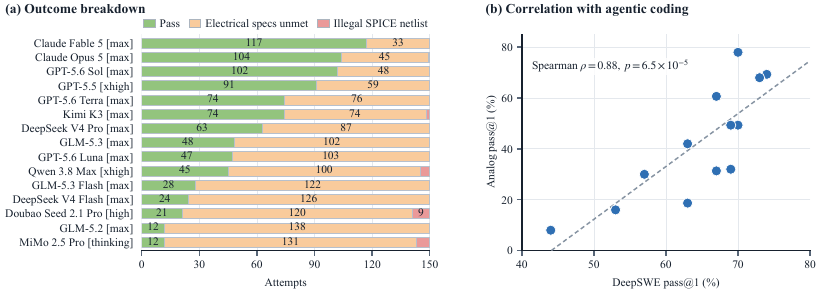}
  \caption{Main-experiment outcomes and coding benchmark scores.
  (a) Outcome breakdown for 15 configurations, with 150 attempts each.
  Pass denotes full electrical acceptance.
  Electrical specs unmet groups all non-passing attempts without a recorded legality rejection, including partial- and zero-reward outcomes.
  Illegal SPICE netlist denotes recorded legality rejection.
  (b) Analog pass rates against DeepSWE leaderboard scores \citep{datacurve2026deepsweleaderboard} for 13 models; the dashed line is a least-squares visual trend, and the inset reports Spearman rank correlation.
  The benchmarks use different protocols, harnesses, and budgets.}
  \label{fig:outcomes-coding}
\end{figure}

The main experiment runs 50 tasks $\times$ 15 configurations $\times$ three rollouts, 2,250 attempts with a two-hour budget each; this section reads off how far current agents get and what they spend (Table~\ref{tab:leaderboard}), and where the difficulty lies (Figure~\ref{fig:outcomes-coding}).

A configuration is a model, a reasoning effort, and a harness. The 15 span the frontier models of Anthropic and OpenAI and the leading models of DeepSeek, Moonshot, Alibaba, Zhipu, Xiaomi, and ByteDance, more than two orders of magnitude apart in cost per attempt; we use the highest available reasoning-effort setting where supported. The four GPT-series and two Claude configurations use their providers' native harnesses, Codex and Claude Code, respectively; the remaining nine use Claude Code through compatible APIs (Section~\ref{sec:levers} measures the harness effect).
Pass@1, the fraction of attempts that satisfy the full contract, is the primary metric; pass@3, the fraction of tasks solved in at least one of three attempts, separates coverage from reliability; R1/R2/R3, the pass rates of the first, second, and third rollouts, expose run-to-run variation; SpecScore, the equal-weight mean of per-attempt gate rewards across tasks and rollouts, diagnoses partial progress.

Table~\ref{tab:leaderboard} shows full-specification pass rates from 8.0\% to 78.0\%.
Claude Fable 5 [max] leads with 117/150 passes, followed by Claude Opus 5 [max] at 69.3\% and GPT-5.6 Sol [max] at 68.0\%.
Reliability and coverage diverge: GPT-5.6 Sol solves 48 tasks at least once, three more than Fable, but solves only 18 in all three attempts versus Fable's 32.
Across the cohort, every task is solved at least once, with task pass rates from 4.4\% to 88.9\%; the suite is neither uniformly unsolved nor saturated.

\begin{table}[H]
  \caption{Main experiment: 150 two-hour attempts per configuration; metrics as defined in the text; cost, output tokens, and turns are per-attempt means. Bold marks the best score and the lowest resource use in each column.}
  \label{tab:leaderboard}
  \vspace{2pt}
  \centering\small
  \setlength{\tabcolsep}{4.5pt}
  \begin{tabular*}{\textwidth}{@{\extracolsep{\fill}} l r r r r r r r}
    \toprule
    & pass@1 & R1\,/\,R2\,/\,R3 & pass@3 & SpecScore & Cost & Out tok & Turns \\
    Configuration & (\%) & (\%) & (\%) & (\%) & (USD) & (k) & \\
    \midrule
    MiMo 2.5 Pro [thinking] & 8.00 & 6\,/\,12\,/\,6 & 16.00 & 22.45 & 2.66 & 332 & 78 \\
GLM-5.2 [max] & 8.00 & 6\,/\,8\,/\,10 & 18.00 & 22.87 & 1.57 & 101 & 39 \\
Doubao Seed 2.1 Pro [high] & 14.00 & 16\,/\,16\,/\,10 & 26.00 & 28.99 & 3.34 & \textbf{63} & 56 \\
DeepSeek V4 Flash [max] & 16.00 & 16\,/\,20\,/\,12 & 24.00 & 30.68 & \textbf{0.12} & 191 & 84 \\
GLM-5.3 Flash [max] & 18.67 & 20\,/\,24\,/\,12 & 38.00 & 33.09 & 0.49 & 382 & 70 \\
Qwen 3.8 Max [xhigh] & 30.00 & 34\,/\,24\,/\,32 & 44.00 & 48.11 & 5.05 & 264 & 85 \\
GPT-5.6 Luna [max] & 31.33 & 28\,/\,32\,/\,34 & 50.00 & 47.53 & 1.39 & 133 & 216 \\
GLM-5.3 [max] & 32.00 & 30\,/\,32\,/\,34 & 52.00 & 45.03 & 2.91 & 237 & 58 \\
DeepSeek V4 Pro [max] & 42.00 & 46\,/\,42\,/\,38 & 62.00 & 55.45 & 0.37 & 293 & 108 \\
Kimi K3 [max] & 49.33 & 52\,/\,48\,/\,48 & 68.00 & 60.87 & 5.52 & 133 & 75 \\
GPT-5.6 Terra [max] & 49.33 & 44\,/\,52\,/\,52 & 72.00 & 68.17 & 8.22 & 115 & 178 \\
GPT-5.5 [xhigh] & 60.67 & 62\,/\,66\,/\,54 & 86.00 & 73.02 & 11.92 & 100 & 133 \\
GPT-5.6 Sol [max] & 68.00 & 72\,/\,62\,/\,70 & \textbf{96.00} & 77.02 & 14.63 & 74 & 148 \\
Claude Opus 5 [max] & 69.33 & 76\,/\,72\,/\,60 & 90.00 & 86.62 & 22.37 & 165 & 46 \\
Claude Fable 5 [max] & \textbf{78.00} & \textbf{80}\,/\,\textbf{74}\,/\,\textbf{80} & 90.00 & \textbf{89.26} & 31.56 & 108 & \textbf{37} \\
\bottomrule
  \end{tabular*}
\end{table}
\vspace{-2pt}

\paragraph{Resource use.}
\label{sec:resources}
Pass rates say what a configuration achieves, not what it spends or how it searches, so Appendix~\ref{app:resources} pairs pass@1 with six per-attempt measures across all 15 configurations: model iterations, tool calls, recorded tool wait, recovered candidate circuits, output tokens, and task runtime.
Pass@1 shows weak rank association with iterations, tool calls, or recovered candidates (Spearman $\rho=+0.10$, $+0.08$, $+0.12$) and a negative association with output tokens ($\rho=-0.28$): GLM-5.3 Flash [max] emits about 3.5 times as many output tokens as the leader despite its lower pass rate.
Across configurations, recorded tool wait rises with pass@1 ($\rho=+0.81$), while mean task runtime falls ($\rho=-0.91$).
Passing attempts last a median of 72.0 minutes versus 118.9 minutes for failures, so configurations with more passes tend to have shorter average runs; this does not mean stopping early causes success.
Section~\ref{sec:case} examines individual search trajectories.

\paragraph{Beyond syntax.}
\label{sec:failures}
\textbf{At the submission boundary, unmet electrical specifications dominate recorded failures.}
A legal SPICE netlist can simulate successfully while implementing the wrong function or missing a required performance target.
Figure~\ref{fig:outcomes-coding}a shows that only 24 of 2,250 attempts (1.1\%) receive a recorded legality rejection; 1,364 of the 1,388 non-passing attempts (98.3\%) have no such rejection but fail electrical acceptance.
This category spans simulation failures, functional failures, and missed performance limits, including partial-reward outcomes: a circuit may earn credit for low power while failing its gain requirement, or function as intended while narrowly missing a single limit.
This endpoint analysis does not classify syntax, testbench, or simulator errors that agents encounter and repair during search.

\paragraph{Beyond coding.}
Figure~\ref{fig:outcomes-coding}b compares our scores with published DeepSWE results~\citep{huang2026deepswe,datacurve2026deepsweleaderboard} for the 13 models with a leaderboard entry; the remaining two have no published result.
Rankings are positively associated (Spearman $\rho=0.88$, $p=6.5\times10^{-5}$, $n=13$), and for the three leading configurations the two scores lie within eight points of each other.
Below the top the two scales diverge: coding scores span 30 points (44\% to 74\%) while analog pass rates span 70 points, and six models with DeepSWE scores between 67 and 70 reach analog pass rates from 31\% to 78\%.
The two benchmarks differ in protocol, harness, and budget, so the gap is not a calibrated transfer loss, but it shows that a coding score is an incomplete proxy for analog-design performance.

\section{Performance Factors}
\label{sec:performance-factors}

\begin{keyfindings}
\begin{itemize}[leftmargin=*]
    \item \textbf{Test-time scaling} through longer runs and greater \textbf{reasoning effort} improves pass rates, with the largest continued gains concentrated at higher effort settings.
    \item \textbf{Harness differences} narrow with time, while \textbf{general-purpose skills} have small or inconsistent effects on pass rates.
    \item \textbf{Task-matched topology references} yield substantial benefits: higher final pass rates or earlier passing solutions.
\end{itemize}
\end{keyfindings}

For selected configurations from the main experiment, we extend each run from two to six hours and record intermediate circuit revisions.
We score completed checkpoint replays and summarize the resulting trajectories on a five-minute grid.
The model and skill studies in panels (a) and (d)--(f) average three attempts per task.
For practical reasons related to cost and changing model availability, the effort and harness studies in panels (b) and (c) use one rollout per task.

\subsection{Test-time scaling}
\label{sec:levers}

\begin{figure}[t]
  \centering
  \includegraphics[width=\linewidth]{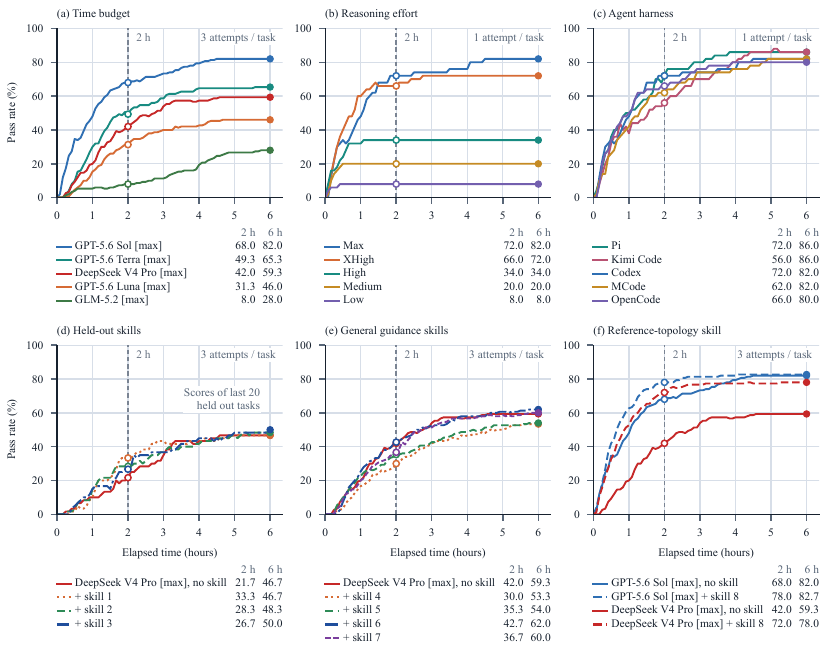}
  \caption{Six-hour pass rates from checkpoint-level scoring, averaged over three attempts per task in panels (a) and (d)--(f). Panels (b) and (c) use one attempt per task. Every point uses each run's latest scored circuit; open and filled markers indicate the two-hour and six-hour scores listed in the key (\%).
  (a)~Five main-experiment configurations.
  (b)~GPT-5.6 Sol in Codex at five reasoning-effort settings.
  (c)~GPT-5.6 Sol [max] in five harnesses.
  (d)~Skills 1--3 on the 20 held-out tasks.
  (e)~Skills 4--7 on the full suite.
  (f)~Reference-topology skill 8 for DeepSeek V4 Pro [max] and GPT-5.6 Sol [max].
  }
  \label{fig:timecourse-summary}
\end{figure}

\paragraph{More time (Figure~\ref{fig:timecourse-summary}a).}
Extending the horizon from two to six hours improves all five shown configurations by 14.0--20.0 percentage points, consistent with the time dependence reported by \citet{zhu2026edgebench}; some trajectories plateau while others continue improving late in the run.

\paragraph{Reasoning effort (Figure~\ref{fig:timecourse-summary}b).}
Holding the base model and Codex harness fixed, we compare five effort settings: Low, Medium, High, XHigh, and Max.
Reasoning effort affects not only final performance but also whether progress continues with additional time. Across the five settings, pass rates range from 8.0--72.0\% at two hours and 8.0--82.0\% at six hours.
Max and XHigh gain a further 10.0 and 6.0 percentage points after two hours, whereas High, Medium, and Low show no additional gains.
Under Low effort, every run that remains unsuccessful at six hours makes its final circuit revision within the first hour, suggesting that additional wall-clock time does not translate into continued circuit exploration for these runs.

\paragraph{Agent harness (Figure~\ref{fig:timecourse-summary}c).}
With GPT-5.6 Sol [max] and the task set fixed, the five harnesses span 56.0--72.0\% at two hours but narrow to 80.0--86.0\% at six hours, corresponding to a maximum difference of only three tasks out of 50.
In contrast, the five model configurations in panel (a) span 54 percentage points at six hours.
Thus, in this experiment, long-horizon performance varies much less across harnesses than across model configurations.

\subsection{Design skills}
\label{sec:skills}

Analog design requires both selecting a circuit topology and sizing it to meet electrical specifications.
To separate these challenges, we vary the knowledge supplied to DeepSeek V4 Pro [max] from textual design guidance to exact task-matched topology blueprints.

\paragraph{General guidance (Figure~\ref{fig:timecourse-summary}d, e).}
General guidance provides little benefit at six hours.
For the held-out study, GPT-5.6 Sol [max] distilled main-experiment trajectories of the first 30 tasks into a handbook (skill 1), a compact workflow (skill 2), and a failure-diagnosis decision tree (skill 3).
DeepSeek V4 Pro [max] received one document at a time on the final 20 tasks, with three attempts per task.
Skill 1 exceeds the baseline by 11.7 percentage points at two hours, yet all three formats finish within 3.3 percentage points of the baseline at six hours.
Because the ordered split changes the family mix, with power management contributing 10 of the 30 source tasks but only 1 of the 20 held-out tasks, this experiment also tests transfer across task families.
On the full suite, three trajectory-distilled documents (skills 4--6) and a task-derived knowledge library (skill 7) finish at 53.3\%, 54.0\%, 62.0\%, and 60.0\%, compared with 59.3\% without a skill.
These four documents reuse information from the evaluated tasks and therefore measure task-informed knowledge reuse.
Skills 4 and 5 trail the baseline by 12.0 and 6.7 percentage points at two hours, while only skills 6 and 7 provide small gains at six hours.

\paragraph{Reference topology (Figure~\ref{fig:timecourse-summary}f).}
Reference-topology guidance produces the largest gain and approximates circuit-IP reuse in engineering practice.
Skill 8 provides one task-matched, non-runnable topology blueprint per task.
Each blueprint preserves device types, connectivity, hierarchy, and interface, while replacing device dimensions, multiplicities, bias ratios, and passive values with placeholders.
This treatment removes topology search while leaving numerical sizing and electrical closure to the agent.
With the reference library supplied, DeepSeek V4 Pro [max] finishes at 78.0\%, adding 28 passing attempts out of 150 and exceeding its baseline by 30.0 percentage points at two hours and 18.7 percentage points at six hours.
The same library supplied to GPT-5.6 Sol [max] raises its pass rate by 10.0 percentage points at two hours but only 0.7 percentage points at six hours.

\paragraph{Interpretation.}
Skills 1--7 provide broadly applicable workflow, diagnostic, or circuit-design guidance, but their limited or inconsistent gains suggest that such textual guidance adds relatively little task-specific information in this setting.
Skill 8 instead provides task-matched circuit structure without solved sizing.
The resulting gains suggest that task-matched reference topologies provide more useful task-specific information than textual design guidance in this setting: topology selection and netlist construction appear to be substantial bottlenecks for DeepSeek V4 Pro [max], while GPT-5.6 Sol [max] primarily benefits from reaching passing solutions sooner.
Exact task-to-reference matching represents an upper bound on practical circuit-IP retrieval, where the closest available design may only approximate the target and require structural adaptation as well as numerical sizing.
A matched topology does not eliminate the nonlinear closure problem: useful sizing changes still depend on the operating point and the active bottleneck~\citep{razavi2001design,gray2009analysis}.%
\section{Case Study: Analog Design Trajectories}
\label{sec:case}

\begin{figure}[!t]
  \centering
  \includegraphics[width=\linewidth]{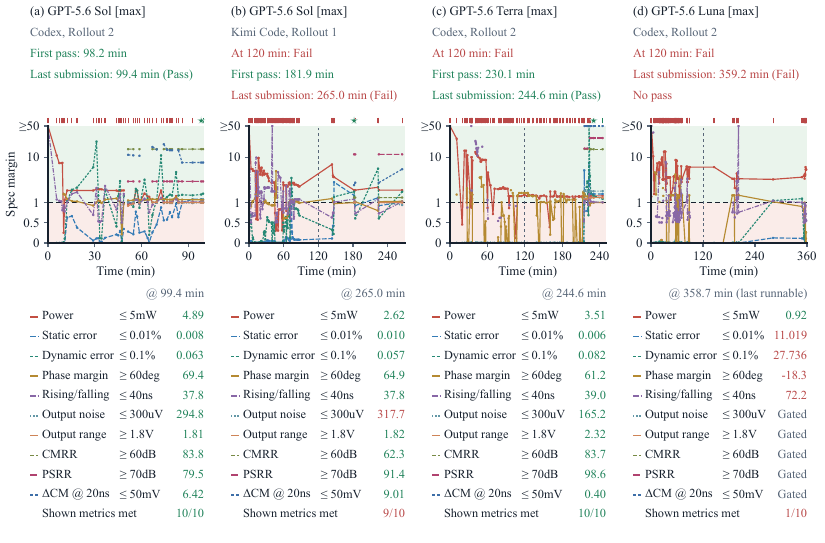}
  \caption{Four six-hour trajectories on one sampling-feedback OTA (task No. 33): early pass (a), pass-then-regress (b), delayed pass (c), and no pass (d).
  Curves show normalized margins for ten metrics drawn from the verifier's 15 scoring gates (higher is better); tables report the last runnable circuit, with blocked metrics marked Gated.}
  \label{fig:case-study}
\end{figure}

Figure~\ref{fig:case-study} follows four six-hour runs on the same fully differential sampling-feedback OTA (task No. 33), whose contract couples settling, accuracy, noise, and common-mode control across process, supply, temperature, and mismatch.
GPT-5.6 Sol [max] in Codex \emph{passes within two hours}: its 98.2-minute circuit passes every scoring gate and its final submission at 99.4 minutes keeps them.
One attempt of GPT-5.6 Sol [max] in Kimi Code \emph{passes, then regresses}: it first passes at 181.9 minutes, keeps editing, and its last submission at 265.0 minutes fails on output noise, 317.7\,$\mu$V against 300\,$\mu$V.
GPT-5.6 Terra [max] \emph{passes after two hours}: failing at 120 minutes, it first passes at 230.1 minutes and its final circuit at 244.6 minutes satisfies every check.
GPT-5.6 Luna [max] \emph{never passes}: its last runnable circuit at 358.7 minutes meets only the power limit, with 11.0\% static error against 0.01\%, and its final submission widens the two output transistors from 1 to 4\,$\mu$m at a multiplicity of 10,000, pushing the transistor model outside its valid range so that the simulator returns no operating point and every check is blocked.
Together, these cases illustrate that nonlinear, coupled specifications make analog search sensitive to topology, sizing, operating point, and measurement coverage, so reliable progress needs physical diagnosis and the agent's own regression testing across process, supply, temperature, and mismatch: the verifier's verdict is hidden, so a run that passes and then regresses cannot know it had passed.%
\section{Discussion and Conclusion}

Our results support three conclusions. First, coding scores are an incomplete proxy for analog design: DeepSWE scores between 67 and 70 accompany analog pass rates from 31\% to 78\%. The difficulty lies in electrical closure, not syntax: recorded legality rejections account for 1.1\% of attempts.
Second, additional time adds 14.0--20.0 points from two to six hours; reasoning effort affects both final performance and continued progress, while harness differences narrow over time.
General and trajectory-distilled skills move endpoints by between $-6.0$ and $+3.3$ points, while supplying the task's reference topology raises DeepSeek V4 Pro [max] by 18.7 points and mainly accelerates GPT-5.6 Sol [max].
Third, across the 15 configurations, pass@1 has weak rank associations with model iterations, tool calls, and recovered candidates ($\rho=+0.10$, $+0.08$, $+0.12$); failing attempts usually run to the two-hour cap.

These findings suggest two directions.
First, agents should retrieve and adapt designs from real circuit libraries: the closer the retrieved design is to the target, the larger the benefit should be, with the exact reference topology as the idealized best case, and the benchmark can measure how much of that benefit realistic retrieval retains.
Second, agents need more reliable closure: checking margins across operating conditions, respecting the valid ranges of device models, and keeping the circuits that pass their own checks.
The released benchmark and trajectories make both measurable.

\paragraph{Limitations.}
The benchmark uses one open-source process design kit, SKY130, at schematic level, so its absolute circuit performance is not comparable to advanced commercial processes, while layout, parasitics, and silicon measurements remain outside scope.
The suite covers only tasks executable with this open toolchain; workflows requiring proprietary models, simulators, or analyses are excluded.
Passing the benchmark does not imply a commercially competitive chip.%
\section*{Reproducibility Statement}

The supplementary material will contain an anonymized release of the benchmark: the 50 task packages with their electrical contracts, starter files, and reference results; the verifier container with the pinned ngspice and SKY130 versions and the electrical benches hidden during the reported experiments; and the eight skill packages with the frozen task partition of the held-out study. The archived trajectories, circuit revisions, and verifier verdicts of every main-experiment and six-hour attempt will be released online.
Appendix~\ref{app:tasks} lists the tasks, Appendix~\ref{app:data-quality} documents the cohort and checkpoint records, Appendix~\ref{app:resources} defines every resource measure, Appendix~\ref{app:skills} describes how each skill was built and evaluated, and Appendix~\ref{app:case} gives the provenance of the case-study trajectories.
Analysis scripts regenerate the results tables and quantitative plots from the frozen data snapshot.
Model sampling is stochastic, so reruns will not reproduce individual trajectories, but every reported number can be recomputed from the archived runs.
Future benchmark versions will introduce held-out tests that remain inaccessible to evaluated agents.%
\section*{Broader Impact and Ethics Statement}

Analog Design Bench improves reproducibility through shared tasks, an open-source toolchain, and specification-based electrical verification, and lowers the barrier to studying agentic analog design without commercial EDA licenses.
Passing the benchmark means meeting the listed electrical specifications, not certifying a circuit for fabrication: unconstrained component values, transistor multiplicity, or area can be impractical (Appendix~\ref{app:evidence-limits}).
The benchmark involves no personal data and no proprietary information: all tasks are built on the open-source SKY130 PDK, and task contributions were provided by the participating designers for release without proprietary circuits, models, or specifications.%
\section*{AI Use Statement}

AI assistance was used to inspect the repository, validate and summarize the frozen rollout table, generate plotting and LaTeX scaffolding, and edit prose.
Quantitative figures use recorded experimental results.
The authors are responsible for verifying all AI-assisted analysis, references, technical claims, illustrations, and the final manuscript.%

\bibliography{references}

\section*{Authors Information}

\begin{description}
  \item[ETH Zurich] Zhishuai Zhang, Xiongjie Zhang, Suyang Song, Seungki Hong, Taekwang Jang
  \item[Tsinghua University] Xiyu He, Kezhuo Liu, Siyu Huang, Yihan Wang,
    Huan Zhang, Minhu Wang, Haikun Jia, Yan Lu, Lu Jie, Nan Sun
  \item[Nanjing University of Posts and Telecommunications] Jing Wang, Yufeng Guo
  \item[Nanjing University] Quanrong Zhuang
  \item[Shanghai Jiao Tong University] Haochi Ying
  \item[Beihang University] Shixuan Wang, Yilin Xu, Zhiguo Tong
  \item[Massachusetts Institute of Technology] Yan Xu
  \item[University of Zurich] Siqi Liu
  \item[Fudan University] Lihan Cui
  \item[California Institute of Technology] Liqun Feng
  \item[The University of Hong Kong] Lei Li
  \item[University of Cambridge] Yao Lai
  \item[MiniMax] Qidi Xu, Pengyu Zhao, Junjie Yan
\end{description}

\beginappendix
\setcounter{figure}{0}
\setcounter{table}{0}
\renewcommand{\thefigure}{A\arabic{figure}}
\renewcommand{\thetable}{A\arabic{table}}

\section{Extended related work}
\label{app:related}

This appendix extends the analog side of Section~\ref{sec:related}; the coding-agent benchmarks that motivate execution-graded evaluation are cited there.

\paragraph{Learning-based analog automation.}
Before LLM-based systems, learned analog automation addressed simulator-guided sizing and task-specific topology generation.
Bayesian optimization~\citep{lyu2018weibo}, AutoCkt~\citep{settaluri2020autockt}, and GCN-RL~\citep{wang2020gcnrl} search parameterized circuits, AnalogGym~\citep{li2024analoggym} packages executable sizing tasks for such optimizers, and CktGNN~\citep{dong2023cktgnn}, LaMAGIC and LaMAGIC2~\citep{chang2024lamagic,chang2025lamagic2}, and AnalogGenie and AnalogGenie-Lite~\citep{gao2025analoggenie,gao2025analoggenielite} generate topologies.
Their action spaces, circuit families, and acceptance criteria are tied to individual automation flows, so they do not measure whether a general-purpose, tool-using agent can complete an unseen design task.

\paragraph{LLM-based analog design.}
LLM-based systems place the model inside the design loop as generator, reasoner, or controller.
Artisan~\citep{chen2024artisan} specializes a domain model for operational-amplifier synthesis, ADO-LLM~\citep{yin2024adollm} combines in-context circuit knowledge with Bayesian optimization, and AnalogCoder and AnalogCoder-Pro~\citep{lai2025analogcoder,lai2026analogcoderpro} generate circuit code and revise it from simulator feedback.
LEDRO~\citep{kochar2025ledro}, White-Box Reasoning~\citep{chen2025whitebox}, TopoSizing~\citep{wei2025toposizing}, and AnaFlow~\citep{ahmadzadeh2025anaflow} make the injected design knowledge explicit through search-space reduction, $g_m/I_D$ grounding, topology-aware annotation, or interpretable sizing flows, and AnalogXpert~\citep{zhang2025analogxpert} and AMSnet-KG~\citep{shi2025amsnetkg} inject circuit-design expertise or knowledge-graph retrieval into topology synthesis.
Agentic systems widen the loop: AmpAgent~\citep{liu2024ampagent} adapts literature designs across processes and targets, Atelier~\citep{shen2026atelier} delegates topology, analysis, sizing, and revision to specialized agents, HeaRT~\citep{poddar2025heart} structures simulator-guided reasoning hierarchically, AnalogAgent~\citep{bao2026analogagent} distills execution feedback into reusable memory, and AnalogTester~\citep{chen2025analogtester} and AnalogVerifier~\citep{liu2026analogverifier} automate testbench construction and verification.
Each system reports results on its own task selection, simulator budget, technology, and verification boundary.

\paragraph{Analog datasets and benchmarks.}
AMSNet~\citep{tao2024amsnet} pairs schematics with netlists; AICircuit~\citep{mehradfar2024aicircuit} supplies simulated analog and RF tasks for performance prediction and specification-to-parameter learning; AMSbench~\citep{shi2025amsbench} and CircuitSense~\citep{akbari2025circuitsense} test multimodal circuit perception and reasoning; SPICEPilot~\citep{vungarala2024spicepilot} grades SPICE program generation; NetlistBench~\citep{ma2026netlistbench} tests structure-preserving netlist recognition and manipulation; and RF-Agent~\citep{xing2026rfagent} evaluates RF-specific reasoning.
Their success criteria are answer accuracy, retrieval quality, prediction error, syntax, structural correctness, or specification satisfaction inside one framework.
Analog Design Bench evaluates the same general-purpose agents on a fixed suite, admits any legal transistor topology, and re-simulates the submitted circuit in an isolated verifier.
The electrical contract is public; only the verifier implementation and its feedback remain inaccessible during an attempt.

\section{Task inventory}
\label{app:tasks}

Table~\ref{tab:tasks} lists the 50 scored tasks with a one-line design objective and the number of condition-level electrical acceptance checks in each contract; the full machine identifiers stay in the benchmark manifest.
Slot numbers label the columns of Figure~\ref{fig:matrix}, and Table~\ref{tab:families} summarizes outcomes by functional family.

\begin{figure}[H]
  \centering
  \includegraphics[width=\linewidth]{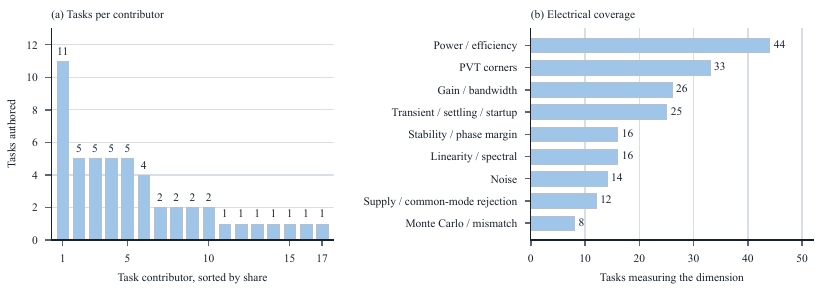}
  \caption{Suite diversity.
  (a)~Tasks authored per contributor; every task has exactly one expert author.
  (b)~Number of tasks whose isolated verifier measures each electrical dimension, classified from the scoring-gate names.}
  \label{fig:suite-diversity}
\end{figure}

\begingroup
\footnotesize
\setlength{\tabcolsep}{3pt}
\begin{longtable}{r>{\raggedright\arraybackslash}p{0.17\linewidth}rr>{\raggedright\arraybackslash}p{0.48\linewidth}}
\caption{The 50 scored design tasks. Solved by is the number of configurations, out of 15, that pass the task at least once across three attempts. Checks gives the condition-level electrical acceptance count under the convention in Appendix~\ref{app:data-quality}.}\label{tab:tasks}\\
\toprule
Slot & Family & Solved by & Checks & Design objective \\
\midrule
\endfirsthead
\toprule
Slot & Family & Solved by & Checks & Design objective \\
\midrule
\endhead
1 & Power management and references & 15 & 30 & Implement a half-bridge class-D power amplifier targeting > 95\% peak efficiency and > 30 mW output power. \\
2 & Power management and references & 15 & 360 & Implement a programmable ICC/IPTAT NMOS current mirror targeting 1.00 mA nominal output current and $\leq$ 8\% output-current error. \\
3 & Signal-chain amplifiers and active filters & 12 & 75 & Implement a source-degenerated GM-R differential amplifier targeting 1.98--2.02 V/V gain and > 60 MHz bandwidth. \\
4 & Signal-chain amplifiers and active filters & 14 & 25 & Implement a source-degenerated GM-R differential amplifier targeting 1.98--2.02 V/V gain and > 70 MHz bandwidth. \\
5 & Interfaces and drivers & 5 & 243 & Implement an all-NMOS half-bridge bootstrap gate driver with a designable off-chip bootstrap capacitor, targeting 3--7 ns dead time and < 3.5 mW driver power. \\
6 & Signal-chain amplifiers and active filters & 9 & 13 & Implement a three-stage switched-capacitor ring amplifier targeting 7.2--8.8 V/V closed-loop gain and $\leq$ 50 ns settling time. \\
7 & Power management and references & 6 & 638 & Implement a PMOS-pass LDO targeting 1.20 V output within $\pm$20 mV and $\leq$ 250 mV dropout. \\
8 & Signal-chain amplifiers and active filters & 3 & 45 & Implement a differential push-pull source-follower buffer targeting > 0.99 V/V gain through 40 MHz and > 90 dB SFDR at 1 MHz. \\
9 & RF, timing, and high-speed & 13 & 65 & Implement a static CML divide-by-2 frequency divider targeting 1--10 GHz input operation and $\geq$ 200 mVpp differential output swing. \\
10 & Signal-chain amplifiers and active filters & 4 & 105 & Implement an optical-receiver transimpedance amplifier targeting > 8 k$\Omega$ transimpedance and > 750 MHz bandwidth. \\
11 & Power management and references & 7 & 361 & Implement a high-impedance first-order bandgap core targeting 1.18--1.26 V output and $\leq$ 50 ppm/$^\circ$C temperature coefficient. \\
12 & Power management and references & 12 & 268 & Implement a self-starting beta-multiplier current reference targeting 40 $\mu$A nominal output current and $\leq$ 8\% compliance variation. \\
13 & Signal-chain amplifiers and active filters & 10 & 352 & Implement a gain-stable differential amplifier targeting 3.0--4.0 V/V differential gain and $\geq$ 30 MHz bandwidth. \\
14 & RF, timing, and high-speed & 8 & 11 & Implement a low-power oscillator targeting 1.8--2.2 MHz output frequency and $\leq$ 20 $\mu$W nominal total power. \\
15 & Power management and references & 9 & 32 & Implement an unregulated charge pump targeting 2.2--2.8 V enabled output and $\leq$ 5 mVpp output ripple. \\
16 & General-purpose op amps and OTAs & 8 & 216 & Implement a two-stage Miller op amp targeting $\geq$ 60 dB open-loop gain and $\geq$ 200 MHz unity-gain bandwidth. \\
17 & Power management and references & 13 & 326 & Implement an NMOS-pass LDO targeting 0.35--0.45 V regulated output and > 45$^\circ$ phase margin. \\
18 & Power management and references & 8 & 29 & Implement a regulated charge pump targeting output within 5\% of 1.3$\times$VDD and $\leq$ 5 mVpp output ripple. \\
19 & Power management and references & 13 & 423 & Implement a two-phase switched-capacitor 2:1 converter targeting a 0.42--0.51 heavy-load conversion ratio and $\geq$ 70\% heavy-load efficiency. \\
20 & Interfaces and drivers & 13 & 2,040 & Implement an 8-to-1 analog multiplexer targeting -0.001--0.001 dB selected-input gain and $\leq$ -80 dB unselected-input gain. \\
21 & Power management and references & 7 & 21 & Implement a high-PSRR bandgap reference targeting 1.05--1.35 V output and $\leq$ 50 ppm/$^\circ$C temperature drift. \\
22 & Data conversion and sampling & 10 & 66 & Implement a differential bootstrap sampler gate driver targeting $\geq$ 72 dB sampled SDR and $\leq$ 500 $\mu$W supply power. \\
23 & Data conversion and sampling & 8 & 2,083 & Implement a 5-bit switched-resistor DAC targeting $\leq$ 0.25 LSB INL and $\leq$ 0.25 LSB DNL. \\
24 & RF, timing, and high-speed & 13 & 597 & Implement a current-starved ring VCO targeting $\geq$ 30 MHz/V minimum tuning gain and $\leq$ 200 $\mu$W supply power. \\
25 & General-purpose op amps and OTAs & 2 & 797 & Implement a complementary rail-to-rail Class-AB operational amplifier targeting $\geq$ 90 dB open-loop gain and $\geq$ 1 MHz unity-gain bandwidth. \\
26 & Signal-chain amplifiers and active filters & 7 & 392 & Implement a fully differential OTA-C biquad targeting a 1.80--2.20 MHz center frequency and a 0.65--0.75 quality factor. \\
27 & Data conversion and sampling & 11 & 110 & Implement a fully differential flash ADC targeting 4-bit resolution and 50 MS/s sample rate. \\
28 & Data conversion and sampling & 2 & 70 & Implement a fully differential floating charge-transfer residue amplifier targeting 5.5--6.5 V/V sampled gain and $\geq$ 60 dB SFDR. \\
29 & General-purpose op amps and OTAs & 8 & 550 & Implement an Ahuja-compensated two-stage OTA targeting $\geq$ 58 dB low-frequency gain and $\geq$ 6.5 MHz unity-gain bandwidth. \\
30 & RF, timing, and high-speed & 7 & 203 & Implement a 2.4 GHz Gilbert-cell mixer targeting $\geq$ 3 dB conversion gain and $\leq$ -45 dB LO-to-IF isolation. \\
31 & Data conversion and sampling & 6 & 47 & Implement a differential bottom-plate sample-and-hold targeting $\leq$ 5 mV acquisition error and $\geq$ 80 dB sampled SFDR. \\
32 & RF, timing, and high-speed & 12 & 93 & Implement a fully differential CML transmitter driver targeting 28 Gb/s NRZ operation and $\geq$ 320 mV eye height. \\
33 & Signal-chain amplifiers and active filters & 4 & 370 & Implement a fully differential gain-1 sampling-feedback OTA targeting $\leq$ 40 ns settling time and $\leq$ 300 $\mu$Vrms output noise. \\
34 & Signal-chain amplifiers and active filters & 2 & 399 & Implement a fully differential gain-8 sampling-feedback OTA targeting $\leq$ 10 ns settling time and $\leq$ 1 mVrms output noise. \\
35 & General-purpose op amps and OTAs & 5 & 267 & Implement a fully differential two-stage Miller op amp targeting $\geq$ 60 dB differential gain and $\geq$ 100 MHz unity-gain bandwidth. \\
36 & General-purpose op amps and OTAs & 7 & 197 & Implement a fully differential telescopic-cascode OTA targeting > 60 dB differential gain and > 50 MHz unity-gain bandwidth. \\
37 & General-purpose op amps and OTAs & 8 & 585 & Implement a three-stage nested-Miller amplifier targeting $\geq$ 110 dB low-frequency gain and $\geq$ 0.4 MHz unity-gain bandwidth. \\
38 & Power management and references & 5 & 218 & Implement a fully integrated external-capacitor-free LDO targeting 1.0 V output within $\pm$30 mV and $\leq$ 100 $\mu$A no-load current. \\
39 & Interfaces and drivers & 5 & 450 & Implement a low-power line driver targeting $\geq$ 0.5 MHz unity-gain bandwidth and $\leq$ 3\% THD at 20 kHz. \\
40 & Data conversion and sampling & 11 & 30 & Implement a first-order 1-bit delta-sigma ADC targeting $\geq$ 40 dB SNDR at OSR 32 and $\geq$ 6 dB/octave noise-shaping improvement. \\
41 & General-purpose op amps and OTAs & 7 & 216 & Implement a current-biased two-stage Miller op amp targeting $\geq$ 70 dB loop gain and $\geq$ 20 MHz unity-gain bandwidth. \\
42 & Data conversion and sampling & 2 & 131 & Implement a fully differential asynchronous SAR ADC targeting 6-bit resolution and 100 MS/s sample rate. \\
43 & RF, timing, and high-speed & 6 & 297 & Implement an inductively degenerated 2.4 GHz LNA targeting $\geq$ 12 dB transducer gain and $\leq$ 2.0 dB noise figure. \\
44 & General-purpose op amps and OTAs & 14 & 144 & Implement a current-biased five-transistor OTA targeting $\geq$ 40 dB loop gain and $\geq$ 100 MHz unity-gain bandwidth. \\
45 & RF, timing, and high-speed & 15 & 13 & Implement a transistor-level divide-by-two flip-flop targeting 500 MHz output frequency and $\leq$ 400 ps rising clock-to-output delay. \\
46 & Data conversion and sampling & 12 & 489 & Implement a 3-bit flash ADC targeting $\leq$ 0.3 LSB static linearity error and 0 dynamic sample errors. \\
47 & RF, timing, and high-speed & 8 & 447 & Implement a PLL charge pump targeting $\leq$ 5\% branch-current error and $\leq$ 2\% UP/DN mismatch. \\
48 & Data conversion and sampling & 5 & 44 & Implement a 4-bit asynchronous SAR ADC targeting > 24 dB SNDR and > 3.90-bit normalized ENOB. \\
49 & RF, timing, and high-speed & 4 & 900 & Implement a differential LC VCO targeting $\geq$ 1.25 tuning ratio and $\leq$ 2.4 mW average power. \\
50 & General-purpose op amps and OTAs & 6 & 240 & Implement a gain-boosted folded-cascode OTA targeting $\geq$ 130 dB DC gain and $\geq$ 200 MHz UGB. \\
\bottomrule
\end{longtable}
\endgroup

\begin{table}[H]
\caption{Descriptive performance by functional family.
The task set is outcome-conditioned, so these values should not be interpreted as intrinsic family difficulty.}
\label{tab:families}
\centering
\footnotesize
\begin{tabularx}{\linewidth}{>{\raggedright\arraybackslash}X rrr}
\toprule
Functional family & Tasks & Full pass (\%) & SpecScore (\%) \\
\midrule
Data conversion and sampling & 9 & 30.4 & 45.5 \\
General-purpose op amps and OTAs & 9 & 28.9 & 41.1 \\
Interfaces and drivers & 3 & 33.3 & 50.4 \\
Power management and references & 11 & 50.7 & 67.9 \\
RF, timing, and high-speed & 9 & 48.9 & 62.2 \\
Signal-chain amplifiers and active filters & 9 & 31.6 & 43.8 \\
\bottomrule
\end{tabularx}
\end{table}

\section{Verification and data quality}
\label{app:data-quality}

\paragraph{Verification boundary and reward hacking.}
Every task exposes its complete electrical contract, circuit interface, allowed device policy, and public development benches.
The independent verifier implements the same disclosed requirements with protected benches and a protected PDK, and returns no feedback during an attempt.
During task conversion, authors and reviewers adversarially inspected trial trajectories for \emph{reward hacking} and revised the packages and legality rules when agents found shortcuts, including attempts to alter PDK temperature coefficients.
At evaluation time only the declared circuit crosses into the verifier sandbox; the legality checker enforces the task's primitive, include, and simulator-directive policy before electrical simulation.

\paragraph{Acceptance-check counting.}
One check is one named electrical criterion at one discrete prescribed condition.
The count expands explicit process, voltage, temperature, load, channel, code-position, fixed-seed mismatch, frequency, polarity, and transition cases.
A lower and upper bound on one scalar form a single interval check.
Raw waveform samples, continuous-sweep points, FFT bins, solver steps, parser guards, and legality checks are excluded, and the resulting checks may be correlated.
The 50 contracts span 11 to 2,083 checks; Table~\ref{tab:tasks} reports every task's count.

\paragraph{Representative expansion.}
The 5-bit switched-resistor DAC in slot 23 illustrates the maximum.
Its public contract specifies INL, DNL, and monotonicity across 32 codes at 11 PVT points and eight fixed-seed mismatch cases, plus endpoint error, power, three checks for each of six directed major-carry transitions, and output resistance at six representative codes.
The semantic expansion is
\[
(11+8)(32+31+31)+11(2+1+6\times3+6)=2{,}083.
\]
At the other extreme, the 2\,MHz oscillator in slot 14 has frequency and output-swing checks at five PVT points plus one nominal-power check, for 11 total.

\paragraph{Simulation failure and pass decision.}
If a required analysis does not converge, times out, or omits a required measurement, the corresponding scoring gate fails or the dependent gates are marked blocked; an absent value never passes.
An early failure can stop later benches, so one submission may execute fewer checks than the complete contract contains.
A submission passes only when it is legal and every scoring gate, including every condition aggregated by that gate, passes.

\paragraph{Limits of the evidence.}
\label{app:evidence-limits}
Task selection is outcome-conditioned: trial runs with six configurations, five evaluated here and one not, were used to exclude easily solved proposals. Consequently, task- and family-level statistics characterize the retained suite rather than intrinsic task difficulty.
Because five of the six selection configurations are also evaluated here, this filtering may make the retained suite disproportionately difficult for these five relative to the other ten evaluated configurations.
The held-out skill split is fixed and uneven, and the full-suite skills reuse the evaluated tasks.
The main experiment and model/skill time courses use three attempts per task, while effort and harness time courses use one; both provide limited stochastic characterization.
The cases in Section~\ref{sec:case} illustrate outcomes and do not estimate their prevalence.
The task scale keeps individual simulations near three minutes; industrial analyses that run for hours or days remain outside the evaluation.
Unless a task constrains them, component area, passive size, and other implementation costs are not acceptance criteria, so a pass does not establish fabrication readiness.
Releasing the verifier supports reproducibility but also exposes it to future agents, creating a potential source of benchmark contamination. Future benchmark versions will therefore use new held-out tests and tasks.

\paragraph{Cohort integrity.}
The main cohort has exactly one row per task, configuration, and rollout, 150 rows per configuration and 45 per task, with no duplicate keys or missing fields; every reward is finite in $[0,1]$, every binary pass equals the indicator $\mathrm{reward}=1$, and every artifact, result, and trajectory path is present.
All 50 tasks have checked-in reference results.
For seven configurations (DeepSeek V4 Flash and Pro [max], GLM-5.2 [max], GPT-5.5 [xhigh], and GPT-5.6 Sol, Terra, and Luna [max]), each reported two-hour result comes from the last circuit revision captured at or before 120 minutes of a six-hour run; the other eight use standalone two-hour runs.
Both groups use the same two-hour scoring cutoff.

\paragraph{Agent harnesses.}
The main experiment runs the four GPT-series configurations in Codex and the other eleven in Claude Code; the harness study of Section~\ref{sec:levers} runs GPT-5.6 Sol [max] in Codex, Kimi Code, MCode, OpenCode, and Pi.
Table~\ref{tab:harnesses} lists the version of each harness.

\begin{table}[H]
\caption{Agent harness versions.}
\label{tab:harnesses}
\centering
\footnotesize
\begin{tabular}{lr}
\toprule
Harness & Version \\
\midrule
Codex & 0.144.1 \\
Claude Code & 2.1.220 \\
Kimi Code & 0.32.0 \\
MCode & 0.2.20 \\
OpenCode & 1.18.18 \\
Pi & 0.80.7 \\
\bottomrule
\end{tabular}
\end{table}

\paragraph{Ideal-element allowances.}
22 tasks permit ideal R and C, 7 tasks permit ideal C, 3 tasks permit ideal R, 2 tasks permit ideal R, C, and L, and 1 task permits ideal R, C, L, and K; the remaining 15 tasks permit no ideal passive elements at all.
\paragraph{Scoring-gate records.}
The grouped scoring-gate count is available for all rows and is distinct from the condition-level check count above.
Within a task, the recorded scoring-gate count differs from the task's modal count in 6 of the 2,250 rows, none of which are full passes; those rows keep their recorded verifier output and are not rescaled.
SpecScore comparisons within the affected tasks should be read up to this residual granularity difference.
\paragraph{Fixed-seed replay.}
Mismatch benches use fixed seeds, so replaying the same circuit under the frozen environment reproduces its verdict.

\paragraph{Checkpoint records.}
The six-hour export indexes 1,750 trajectories and 83,388 scored circuit revisions; 13,041 captured revisions still await replay.
Curves use a five-minute grid, carry the last verified score forward across pending revisions, keep fixed denominators, and count a trajectory once when it belongs to several comparisons.
Among the 990 identity-audited runs with complete checkpoint scores, 15 runs on 12 tasks pass at an earlier checkpoint and then submit a changed circuit that fails final verification, the pass-then-regress pattern of Section~\ref{sec:case}.

\section{Resource accounting}
\label{app:resources}

\begin{figure}[H]
  \centering
  \includegraphics[width=\linewidth]{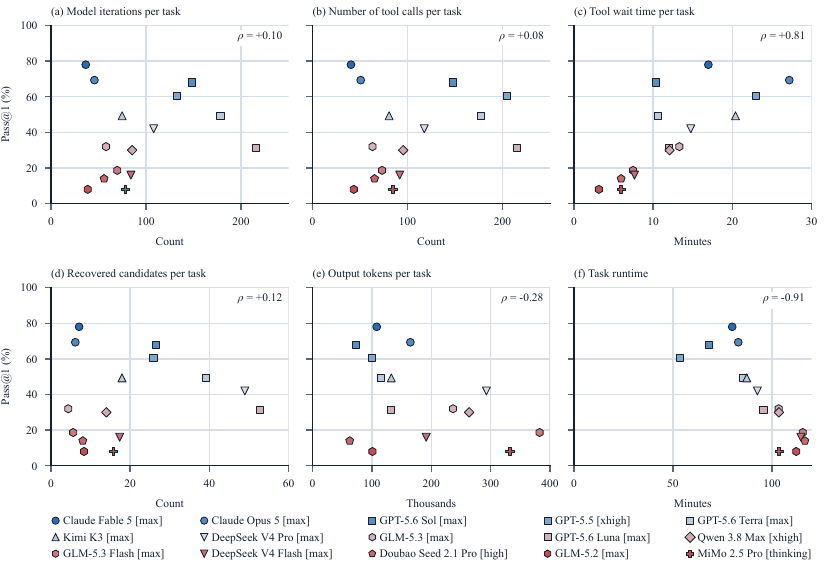}
  \caption{Resource and candidate survey.
  Each point pairs pass@1 with mean resources per task attempt (50 tasks, three rollouts each).
  $\rho$ denotes Spearman correlation.
  Tool wait includes harness overhead and only recorded intervals.
  All panels cover 15 configurations; (d) reports distinct recovered non-template circuit contents within two hours.
  (e) reports output tokens, consistent with Table~\ref{tab:leaderboard}.
}
  \label{fig:cost-score}
\end{figure}

\begin{figure}[H]
  \centering
  \includegraphics[width=\linewidth]{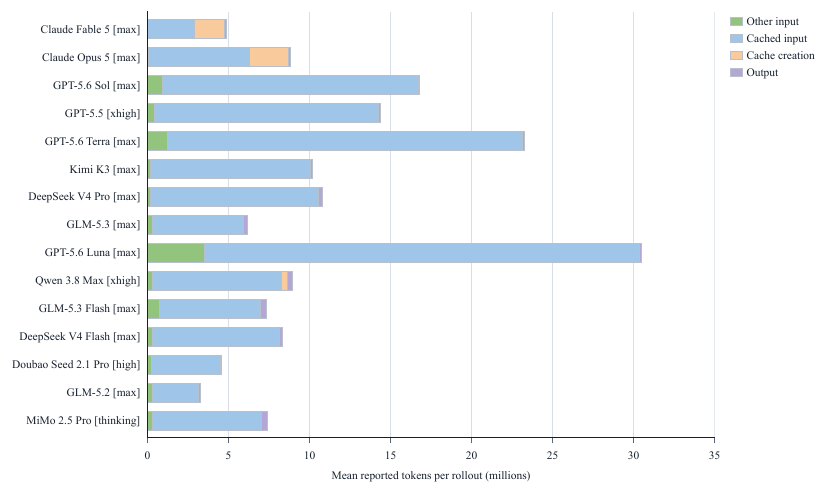}
  \caption{Mean token composition per rollout under the common accounting schema.
  Cached input and cache creation are separated from uncached input without double-counting.}
  \label{fig:usage-breakdown}
\end{figure}

Figure~\ref{fig:cost-score} pairs pass@1 with six per-attempt measures for all 15 configurations of the main experiment; each measure is a mean over the 150 attempts of a configuration, and Figure~\ref{fig:usage-breakdown} splits each configuration's mean token total by category.

\paragraph{Iterations, tool calls, runtime, and tokens.}
Model iterations count recorded model requests.
Tool calls are recounted from the published trajectories and match every attempt's released count; each model-facing invocation counts once, including failed and polling calls, and a wrapper call counts as one event regardless of its internal operations.
Task runtime is the released span from first request to last response.
Tokens and estimated costs use the accounting schema of Table~\ref{tab:leaderboard}, which separates uncached input, cached input, cache creation, and output without double counting.

\paragraph{Tool wait.}
Each matched tool call contributes the interval from the model response that emitted it to the first later request containing its result; overlapping intervals are merged and recorded model response time is subtracted, so parallel calls share elapsed time.
The interval includes simulator execution, dispatch, and harness overhead.
Boundaries are recovered for 99.2\% of the 232,743 tool calls; terminal calls without a following request, unmatched results, and reversed timestamps contribute zero rather than an extrapolated value, so the plotted wait is a lower bound on actual execution time.

\paragraph{Candidate circuits.}
Panel (d) counts distinct task-level circuit contents recorded within two hours, deduplicated by SHA-256 within each attempt and excluding the initial template and any return to it; repeated submissions of identical contents count once, and simulator launches are not counted, since one candidate may be simulated many times and one tool call may launch a batch.
Two sources cover disjoint configurations: trace reconstruction from the public trajectories for eight configurations (1,200 attempts, of which 836 have complete histories and 364 keep lower bounds), and retained \texttt{circuit.spi} snapshots for the other seven (1,050 attempts), with selected submissions checked against their published artifacts where snapshot identity is available.
Together they recover 44,695 per-attempt candidate contents.
Retained snapshots can miss transient or scratch-file contents, while trace reconstruction follows recoverable file changes, so panel (d) describes recorded candidate activity under these capture methods rather than complete histories.

\section{Task-level outcomes}
\label{app:matrix}

\begin{figure}[H]
  \centering
  \includegraphics[width=\linewidth]{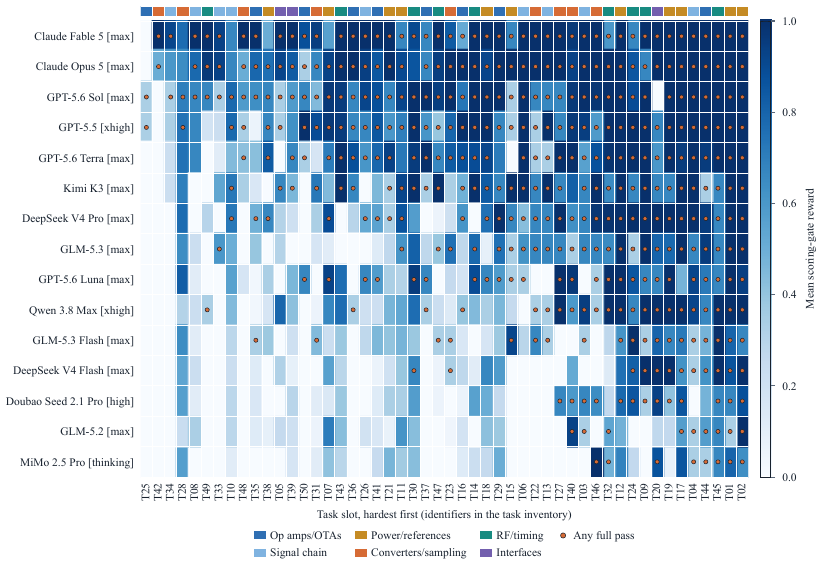}
  \caption{Mean reward for every task/configuration pair in the main cohort.
  Cell shade is the three-rollout mean scoring-gate reward; a dot marks at least one full-spec pass.
  The strip above the matrix gives the functional family of each task slot.}
  \label{fig:matrix}
\end{figure}

Figure~\ref{fig:matrix} shows the three-rollout mean reward for each task--configuration pair, with task columns ordered by the number of configurations solving the task (fewest first; ties by mean reward) and labeled by slot (Table~\ref{tab:tasks}), and configuration rows ordered by leaderboard rank.

\section{Design skills}
\label{app:skills}

Section~\ref{sec:skills} evaluates eight skill packages, each made available to the agent for consultation during an attempt; no attempt receives more than one package, and every treatment runs under the six-hour budget with a no-skill baseline on the same tasks.
Table~\ref{tab:skills} lists the packages and the frozen task partition of the held-out study.
All eight skill packages will be released online.

\begin{table}[t]
\caption{The eight skill packages of Section~\ref{sec:skills}, grouped by how they were built. Every package is evaluated with DeepSeek V4 Pro [max] under the six-hour budget, three attempts per task; skill 8 is also evaluated with GPT-5.6 Sol [max].}
\label{tab:skills}
\centering
\footnotesize
\begin{tabularx}{\linewidth}{c >{\raggedright\arraybackslash}X r}
\toprule
Skill & Package & Size \\
\midrule
\multicolumn{3}{>{\raggedright\arraybackslash}p{\dimexpr\linewidth-4\tabcolsep}}{\emph{Held-out guidance}: distilled by GPT-5.6 Sol [max] from the 1,260 main-experiment trajectories of tasks 1 to 30; evaluated on the held-out tasks 31 to 50} \\
1 & General handbook & 1,295 words \\
2 & Compact workflow & 668 words \\
3 & Failure-diagnosis decision tree & 1,258 words \\
\midrule
\multicolumn{3}{>{\raggedright\arraybackslash}p{\dimexpr\linewidth-4\tabcolsep}}{\emph{Full-suite design documents}: distilled by GPT-5.6 Sol [max] from every successful and failed main-experiment trajectory then available, versions 2 and 3 revising version 1; evaluated on all 50 tasks} \\
4 & Design document, version 1 & 5,716 words \\
5 & Design document, version 2 & 1,372 words \\
6 & Design document, version 3 & 1,042 words \\
\midrule
\multicolumn{3}{>{\raggedright\arraybackslash}p{\dimexpr\linewidth-4\tabcolsep}}{\emph{Circuit knowledge}: a library organized from the 50 task instructions, with no access to solutions, tests, or trajectories; evaluated on all 50 tasks} \\
7 & Topic index, 50 circuit-topic guides, and two cross-cutting guides & 14,490 words \\
\midrule
\multicolumn{3}{>{\raggedright\arraybackslash}p{\dimexpr\linewidth-4\tabcolsep}}{\emph{Reference topologies}: one topology blueprint per task that preserves device types, connectivity, hierarchy, and interface while replacing device dimensions, multiplicities, bias ratios, and passive values with invalid tuning placeholders; evaluated on all 50 tasks} \\
8 & Reference-topology library, one netlist per task & 50 netlists \\
\bottomrule
\end{tabularx}
\end{table}

\paragraph{Skills 1 to 3: held-out guidance.}
The source population is the 1,260 main-experiment trajectories of the first 30 tasks in the original study order: 30 tasks, the 14 configurations evaluated at the time, and three rollouts, comprising 497 full passes and 763 non-passing attempts.
This distillation source predates the cohort reported here: of its 14 configurations, 11 match the reported ones, GPT-5.5 [xhigh] and Doubao Seed 2.1 Pro [high] come from earlier cohorts, and one is not reported in this paper.
GPT-5.6 Sol [max] distilled general process observations from both outcomes, not task-specific circuits, into a general handbook (skill 1), a compact workflow (skill 2), and a failure-diagnosis decision tree (skill 3).
The split follows the original ordered task manifest, slots 1 to 30 as sources and 31 to 50 as held out; it is neither random nor ranked by difficulty, and the extraction did not read the held-out tasks' instructions, trajectories, solutions, or verifiers.
Each document was evaluated with DeepSeek V4 Pro [max] on the 20 held-out tasks, three attempts per task, so each treatment has 60 results.

\paragraph{Skills 4 to 6: full-suite design documents.}
Skills 4, 5, and 6 are three revisions of one design document: GPT-5.6 Sol [max] read every successful and failed main-experiment trajectory then available and distilled the design expertise into a 5,716-word playbook covering budgeting, netlist construction, diagnosis, and final verification (skill 4); skill 5 revises it into a 1,372-word design loop with iteration and stopping rules, and skill 6 into a 1,042-word decision policy that routes each measured failure to the next experiment.
Each revision was evaluated with DeepSeek V4 Pro [max] on all 50 tasks, three attempts per task.

\paragraph{Skill 7: circuit-knowledge library.}
Skill 7 was organized from the 50 task instructions with no access to solutions, tests, trajectories, scores, or artifacts: a topic index, 50 circuit-topic guides on operating principles, topology trade-offs, first-order budgets, and diagnosis, and two cross-cutting guides on devices and budgeting and on simulation and robustness, 14,490 words in all.
Because its topics come from the evaluated tasks' instructions, it measures task-informed knowledge reuse rather than held-out generalization; with it, DeepSeek V4 Pro [max] finishes at 60.0\% against 59.3\% without a skill, one additional passing attempt out of 150, with 11 tasks improving and 12 worsening.

\paragraph{Leakage control and interpretation.}
Skills 1 to 7 deliberately exclude the tasks' reference topologies and solution netlists.
This prevents direct architecture leakage, while limiting the packages to general workflows, failure patterns, and circuit principles that strong pretrained models may already know.
Their small or negative endpoint changes therefore measure the limited value of additional general guidance in this setting.

\paragraph{Skill 8: reference-topology library.}
Skill 8 is a library of 50 task-matched topology blueprints, one per task.
Each preserves the reference device types, connectivity, hierarchy, and interface while replacing every device dimension, multiplicity, bias ratio, and passive value with an invalid \texttt{<TUNE>} placeholder.
The supplied blueprint cannot run until the agent determines all numerical values and closes the electrical specifications.
It was evaluated with GPT-5.6 Sol [max] and DeepSeek V4 Pro [max] on all 50 tasks, three attempts per task.
Because retrieval is exact by construction, the result measures topology reuse under an idealized retrieval condition similar to circuit-IP reuse in engineering practice; held-out transfer is untested.

\section{Case-study evidence}
\label{app:case}

\begin{figure}[t]
  \centering
  \includegraphics[width=\linewidth]{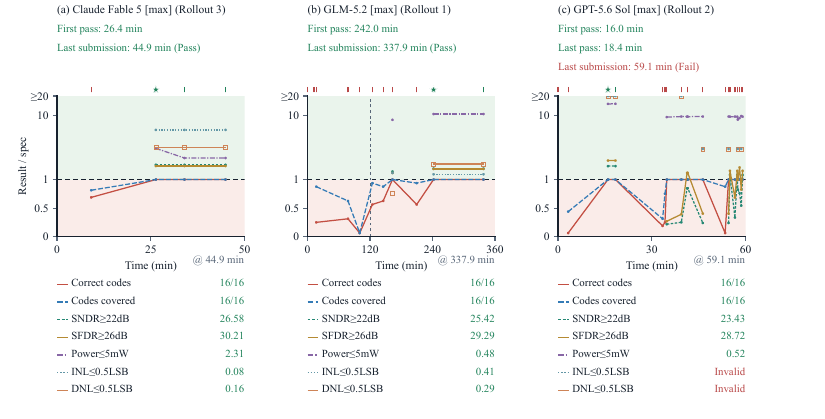}
  \caption{A second case, the 4-bit, 50\,MS/s Flash ADC (task No. 27), showing early closure (a), delayed closure (b), and pass-then-regress (c).
  Rows are direction-aligned result-to-specification ratios of the seven displayed quantities, capped at 20; tables list the final submission.}
  \label{fig:case-flash}
\end{figure}

Figure~\ref{fig:case-study} follows four six-hour runs on task No. 33, the fully differential sampling-feedback OTA: GPT-5.6 Sol [max] in Codex, rollout 2, with 43 retained circuits; GPT-5.6 Sol [max] in Kimi Code, rollout 1, with 84; GPT-5.6 Terra [max] in Codex, rollout 2, with 127; and GPT-5.6 Luna [max] in Codex, rollout 2, with 130.
Figure~\ref{fig:case-flash} follows three six-hour runs on task No. 27, the 4-bit, 50\,MS/s Flash ADC, whose 16-code transfer check gates every downstream measurement: Claude Fable 5 [max], rollout 3 (early closure); GLM-5.2 [max], rollout 1 (delayed closure); and GPT-5.6 Sol [max], rollout 2 (pass-then-regress). Fable and GLM run in Claude Code, and Sol in Codex.
Both selections illustrate distinct outcomes; they do not estimate their prevalence or separate model from harness effects, and the case results are kept apart from the main experiment's scores.

\paragraph{Metric recovery for the OTA.}
All 384 retained circuits were replayed through the frozen verifier image in independent, network-disabled containers with a 1,800-second limit; the verifier's 15 scoring gates and their acceptance rules are unchanged.
Most checkpoints complete only the nominal scoring gate, so each plotted quantity is the worst value among the corners that checkpoint finished; on the 24 checkpoints with a complete metric matrix it equals the verifier's own gate value.
Four replays timed out, all in the Terra run, and are kept as timeout records; missing measurements stay missing, never zero.

\paragraph{Circuit histories for the Flash ADC.}
The panels contain 4, 11, and 20 circuit checkpoints.
Fable's intermediate measurements come from retrospective re-evaluation of reconstructed circuits, and its timestamps locate recorded edits rather than the receipt of passing feedback; GLM and Sol use retained verifier reports.
Fable's 9.1-minute design fails; its 26.4-minute design passes after changes to comparator sizing, integration capacitance, buffering, and the SR latch; its 34.0-minute design changes the reference ladder and tap capacitances and also passes; later edits only change comments, so the panel keeps the three distinct designs and the final file.
GLM first passes at 242.0 minutes and its last submission at 337.9 minutes still passes.
Sol first passes at 16.0 minutes and last at 18.4; its last submission at 59.1 minutes passes the transfer, SNDR, SFDR, and power checks, but a non-monotonic fine-ramp response invalidates its INL and DNL.

\paragraph{Display.}
In Figure~\ref{fig:case-study} the 15 scoring gates reduce to ten rows: power, static error, dynamic error, phase margin, settling (the worse of rising and falling), output noise, output range, CMRR, PSRR (the worse of the two supply polarities), and the common-mode residual 20\,ns after the step; the output common-mode, common-mode peak-deviation, and 100\,ns residual gates are not drawn.
In Figure~\ref{fig:case-flash} the verifier's six scoring gates give seven rows, because transfer contributes both the correct-output fraction and the distinct-code coverage out of 16, and the downstream gates are blocked when transfer is invalid.
Each row is a direction-aligned ratio to its specification, $x/s$ for a lower bound, $s/x$ for an upper bound, and $10^{(x-s)/20}$ for a decibel lower bound, so a ratio of one or more satisfies the check; the axis is linear from zero to one and logarithmic above, with ratios at or beyond the top tick (50 for the OTA, 20 for the Flash ADC) drawn there.
Headers give the first passing circuit and the last submission; each table lists the last circuit that ran, marked last runnable when the final submission did not run, with checks blocked by an earlier failure marked Gated, unevaluated ones N/A, and timed-out replays Timeout.

\FloatBarrier
\end{document}